\documentclass{article}

\usepackage{microtype}
\usepackage{graphicx}
\usepackage{subfigure}
\usepackage{booktabs} 
\usepackage{enumitem}

\DeclareUnicodeCharacter{0301}{\'{}} 
\DeclareUnicodeCharacter{0327}{\c{}}

\usepackage{hyperref}

\usepackage[accepted]{icml2025}
\usepackage{amsmath}
\usepackage{amssymb}
\usepackage{mathtools}
\usepackage{amsthm}

\usepackage[capitalize,noabbrev]{cleveref}

\theoremstyle{plain}

\theoremstyle{definition}

\theoremstyle{remark}

\usepackage[textsize=tiny]{todonotes}

\icmltitlerunning{Preprint}
\usepackage{indentfirst}   
\begin{document}

\twocolumn[
\icmltitle{Convolution-Based Converter : A Weak-Prior Approach For Modeling Stochastic Processes Based On Conditional Density Estimation }



\icmlsetsymbol{equal}{*}

\begin{icmlauthorlist}
\icmlauthor{Chaoran Pang}{yyy,comp}
\icmlauthor{Shuangrong Liu}{yyy}
\icmlauthor{Shikun Tian}{yyy,comp}
\icmlauthor{WenHao Yue}{yyy}
\icmlauthor{Xingshen Zhang}{yyy}
\icmlauthor{Lin Wang}{yyy,comp}
\icmlauthor{Bo Yang}{comp,yyy}
\end{icmlauthorlist}

\icmlaffiliation{yyy}{Shandong Key Laboratory of Ubiquitous Intelligent Computing, University of Jinan, Jinan 250022, China}
\icmlaffiliation{comp}{Quan Cheng Laboratory, Jinan 250100, China}

\icmlcorrespondingauthor{Lin Wang}{wangplanet@gmail.com}
\icmlcorrespondingauthor{Bo Yang}{yangbo@ujn.edu.cn}

\icmlkeywords{Machine Learning, ICML}

\vskip 0.3in
]



\printAffiliationsAndNotice{}  

\begin{abstract}
In this paper, a Convolution-Based Converter (CBC) is proposed to develop a methodology for removing the strong or fixed priors in estimating the probability distribution of targets based on observations in the stochastic process. Traditional approaches, e.g., Markov-based and Gaussian process-based methods, typically leverage observations to estimate targets based on strong or fixed priors (such as Markov properties or Gaussian prior). However, the effectiveness of these methods depends on how well their prior assumptions align with the characteristics of the problem. When the assumed priors are not satisfied, these approaches may perform poorly or even become unusable. To overcome the above limitation, we introduce the Convolution-Based converter (CBC), which implicitly estimates the conditional probability distribution of targets without strong or fixed priors, and directly outputs the expected trajectory of the stochastic process that satisfies the constraints from observations. This approach reduces the dependence on priors, enhancing flexibility and adaptability in modeling stochastic processes when addressing different problems. Experimental results demonstrate that our method outperforms existing baselines across multiple metrics.
\end{abstract}

\section{Introduction}
\label{submission} Stochastic processes provide a powerful mathematical framework for describing dynamic systems that evolve over time under uncertainty 
\cite{doob1942stochastic}, enabling the analysis and prediction of complex dependencies and dynamic behaviors. They find broad applications in diverse real-world problems where uncertainty, variability, and randomness are inherent, such as in financial markets risk analysis \cite{dupacova2002stochastic,jiang2019multifractal}, biological processes \cite{reid1953stochastic}, the design of optimization algorithms \cite{najim2004stochastic}, and spatial data modeling \cite{banerjee2008gaussian}.

Researchers have developed various methods \cite{pavliotis2014stochastic} for modeling stochastic processes. Among these, Stochastic Differential Equations (SDEs)-based methods \cite{van1976stochastic}, known for their simplicity and ease of use, are widely employed to address continuous-time stochastic modeling problems \cite{mao2013stabilization}, by explicitly defining differential equations based on expert knowledge. However, SDE-based methods simplify problem constraints when defining the equation structure and diffusion term, which limits their ability to model complex systems. Additionally, solving SDEs poses significant numerical challenges \cite{wilkie2004numerical}, particularly in high-dimensional or highly nonlinear settings \cite{platen1999introduction}.

Conditional Density Estimation (CDE) methods, e.g., Markov-based models \cite{feller1991introduction} and Gaussian Processes (GPs) \cite{seeger2004gaussian}, can directly capture dependency relationships from data in modeling stochastic processes, without the need to explicitly construct dynamic equations. Markov-based models offer a computationally lightweight framework for modeling sequential dependencies by the assumption that future states depend solely on the current state \cite{chung1967markov}. However, this assumption limits their ability to capture long-term dependencies and complex dynamics. Additionally, these models often struggle to achieve satisfactory performance with limited data due to their high data requirements for accurately representing state transitions \cite{rabiner1989tutorial}. Compared to Markov-based models, GPs introduce a strong prior assumption, where any finite set of random variables follows a joint Gaussian distribution \cite{seeger2004gaussian}. This prior enables GPs to capture complex dependencies among variables and make high-confidence probabilistic predictions, even in dynamic and data-scarce scenarios \cite{wilson2011gaussian}. 

While GPs have demonstrated considerable success across various applications, the strong prior assumption limits their broader applicability \cite{bishop2006pattern}. When the data conforms to the assumed prior, these models can deliver robust performance. However, significant deviations from this prior often lead to substantial performance degradation. 

Neural Network-based CDE methods,such as Mixture Density Network \cite{bishop1994mixture} and Deconvolutional Density Network \cite{chen2022deconvolutional}, aim to adaptively model stochastic process, eliminating the need for strong prior assumptions regarding the explicit structural design of stochastic processes. However, their performance heavily depends on the size of the training data, leading to a substantial reduction in the reliability and generalization performance when faced with scenarios involving limited data\cite{starkman2023stream}. Therefore, developing methods for limited data stochastic process modeling under weak prior conditions remains a critical open challenge.

To address the above challenges, we propose Convolution-Based Converter (CBC), a weak-prior assumption method for establishing stochastic processes that can adapt to adverse situations even with limited data. Unlike above approaches, CBC does not rely on strong or fixed prior assumption, e.g., Markov properties or Gaussian prior, in modeling stochastic process. Instead, it adaptively estimate the probability distribution of random variables and the dependency relationships between of them, through a Convolution-Based Converter.  CBC transforms trajectories of an arbitrary initial stochastic process into trajectories of an expected stochastic process that satisfy observational conditions. By influencing the observations, the entire stochastic process is affected, enabling CBC to maintain  generalization capability even in limited-data scenarios.

The main contributions of this study are summarized as follows. 
\begin{itemize}[leftmargin=1.5em,itemsep=0pt,topsep=0pt]
\item The Convolution-Based converter is proposed to enhance the broad-spectrum adaptability of stochastic process modeling methods across diverse problems by removing strong prior assumptions.

\item The dependence network generation paradigm is designed to model dependencies among random variables in stochastic processes through convolutional-deconvolutional operations.

\item Experiments show that CBC outperforms different types of  stochastic process modeling methods, including strong prior and neural network-based approaches, on diverse problems.
\end{itemize}
\section{Related Work}

\begin{figure*}[htbp]
  \centering
  \includegraphics[width=0.9\textwidth]{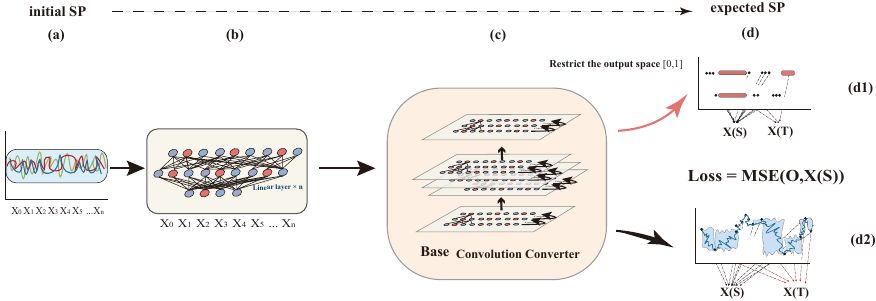}
  \caption{The framework of Convolution-Based Converter (CBC) consists of four successive parts:(a) a input initial stochastic process ,(b) a constructor that establishes preliminary dependency  (c) A Convolution-Based Converter that constructs the dependencies among random variables in the stochastic process (d) The output expected stochastic process. The modeling effect is shown in (d2), where (d1) represents restricting the output space to the [0,1] range.}
  \label{fig:model}
\end{figure*}
\noindent \textbf{SDE-based Models}   Stochastic Differential Equations-based methods (SDEs) are classical approaches for modeling stochastic processes. They utilize explicit differential equations, comprising a drift term and a diffusion term, to describe the instantaneous changes of a stochastic process \cite{oksendal2013stochastic}\cite{kloeden1992stochastic}.
Researchers have developed various modeling strategies based on the SDEs framework, including linear \cite{arminger1986linear}, nonlinear \cite{overgaard2005non}, and jump-diffusion SDEs \cite{jiang2019multifractal}. Key techniques, such as parameter estimation \cite{nielsen2000parameter}, numerical solutions \cite{burrage2004numerical}, and stochastic control \cite{nisio2015stochastic}, have also been extensively explored. These methodological advancements have established SDEs as a fundamental and versatile tool in stochastic process modeling. However, while the theory of SDEs is elegant and capable of describing a wide range of stochastic processes, SDE-based methods oversimplify the real problems by assumptions about system dynamics and noise distributions \cite{przybylowicz2022efficient}. These assumptions are often simplified to ensure analytical tractability, potentially leading to significant deviations from real-world complexities \cite{oksendal2013stochastic}, thus limiting the model's accuracy and generalization ability. Furthermore, SDEs often face difficult-to-solve problems, especially for high-dimensional or nonlinear systems \cite{platen1999introduction}.

\noindent \textbf{Markov-based Models}   Markov-based models, such as Markov chains 
 \cite{chung1967markov} and Hidden Markov Models (HMMs) \cite{eddy1996hidden}, explicitly model state transition probabilities, effectively capturing the dynamic evolution of stochastic processes over time and thus modeling processes with temporal properties. Discrete-time Markov chains \cite{gomez2010discrete} directly represent the conditional probabilities of state transitions at each time step, offering a tractable method for modeling sequence-dependent stochastic processes \cite{craig2002estimation}, typically expressed as $P(X_{i+1}|X_i)$. HMMs further extend this capability by introducing hidden states, allowing the model to capture more complex observation sequences in which observable data are generated based on the evolution of unobserved hidden states.
However, Markov-based models assume that future states depend solely on the current state, making it difficult to capture long-range dependencies and complex dynamics 
\cite{rabiner1989tutorial}. Moreover, the accuracy of Markov models heavily relies on having sufficient data. When data are limited, the estimation of state transition probabilities may become unreliable, thereby affecting both predictive performance and generalization.

\noindent \textbf{Gaussian Processes-based Models}   Gaussian processes (GPs) serve as a conditional probability estimation method by assuming that every finite subset of random variables in the stochastic process follows a multivariate Gaussian distribution 
 \cite{seeger2004gaussian}, we can write a Gaussian process as $GP \sim (m(i),K(i,i'))$, where $m$ is the mean function and $K$ is covariance function, 
Gaussian Process stands as the a CDE approach for modeling stochastic process. It operates under the assumption that each random variable distribution is Gaussian, employing carefully chosen mean and covariance functions to define suitable Gaussian priors for data fitting \cite{williams1995gaussian}. However, this strong reliance on Gaussian assumptions restricts the model’s applicability primarily to data that closely follow Gaussian distributions, thereby severely limiting its flexibility. To address these limitations, various refinements have been introduced, such as Warped Gaussian Processes (WGP) 
\cite{lazaro2012bayesian} that nonlinearly transform Gaussian distributions using activation functions, and t-processes \cite{shah2014student} as well as their variants \cite{jylanki2011robust} \cite{chen2020multivariate} that replace Gaussian distributions with Student-t distributions. Despite these advancements, none has fully circumvented the inherent dependence on specific prior distributional forms.

\noindent \textbf{Neural network–based CDE methods}   Neural network–based CDE methods offer a flexible alternative for modeling stochastic processes by directly learning conditional distributions from data. This approach circumvents the strong prior assumptions typical of Gaussian processes, enabling the modeling of a broader range of distribution shapes. Representative frameworks include Deconvolutional Density Networks (DDN) \cite{chen2022deconvolutional} and Mixture Density Networks (MDN) \cite{bishop1994mixture}, which can capture complex multimodal, heavy-tailed, or other behaviors. However, these methods often require large training datasets and careful network design or hyperparameter tuning.
\begin{figure*}[htbp]
  \centering
  \includegraphics[width=0.8\textwidth]{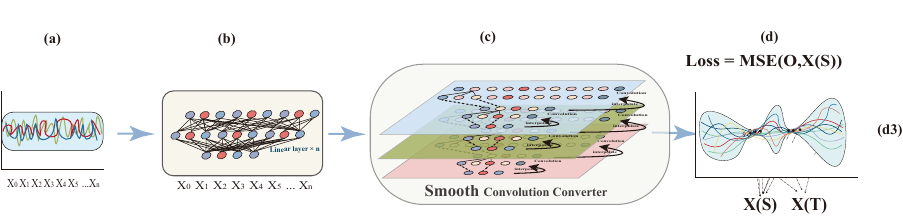}
  \caption{Smooth Convolution-Converter:the framework is as the same as the Fig\ref{fig:model},the difference is replacing the base Convolution-Converter with a Smooth Convolution-Converter to accommodate the modeling of smooth stochastic processes,and d(3) is the expected stochastic process}
  
  \label{fig:Deconv}
\end{figure*}

\section{Methodology}

\subsection{Formulation}
A stochastic process can be represented as $X = \{X(i)\}_{i\in I}$ 
 \cite{doob1942stochastic},  where i serves as the index indicating the locations of the random variables. The index set $I$ is composed of both the observed locations $S$ and the target locations $T$, formally expressed as $I = S \cup T$, for $\forall s \in S$, the corresponding observed values are denoted as $O_s$, where $O=\{O_s|s\in S\} =X(s)$. Let $P$ be the joint probability distribution over the stochastic process,and therefore a conditional distribution $P(X(T)|X(S)=O)$, our task is to predict the conditional probability distribution for every $t \in T$ given O.

General CDEs-based methods often rely on strong priors over the stochastic process \cite{wilson2011gaussian} \cite{rabiner1989tutorial}. However, these strong priors restrict the model's flexibility and can lead to significant deviations when the prior is misspecified. Neural Network-based CDE methods has the potential to weaken such strong priors, However, their performance is highly dependent on large-scale training data, often resulting in compromised performance in limited-data scenarios. Consequently, modeling stochastic processes with weak priors while maintaining reliability in limited-data scenarios remains a critical research challenge.

We introduce a Convolution-Based Converter (CBC), designed to implicitly estimate the conditional probability distribution $P(X(T)|X(S)=O)$ without relying on strong or fixed prior assumptions. Instead, it transforms initial stochastic process trajectories into expected trajectories that are consistent with observations. CBC trains a neural network-based converter to learn how to transform initial stochastic process trajectories into target stochastic process trajectories that satisfy observational constraints. During training, a loss function is applied at the observation points to guide the neural network in continuously optimizing the trajectory transformation. This process ultimately achieves an effective mapping from the initial stochastic process to the stochastic process based on observations, thereby implicitly modeling the stochastic process.

CBC does not impose strong or fixed priors. Instead, it strives to achieve greater flexibility and adaptability within a weak prior framework, allowing for better handling of the diverse challenges in stochastic process modeling. During the transition, we directly generate complete trajectories consisting of both observation and target variables. Although the constraints only affect the observation variables, the entire trajectory is influenced by constructing dependencies among the random variables. This allows the targets to be indirectly estimated through the constraints imposed on observations.Compared to directly modeling $P(X(I)|I)$ can generalize to the entire stochastic process modeling with limited data.

\begin{figure*}[htbp]
  \centering
  \includegraphics[width=0.8\textwidth]{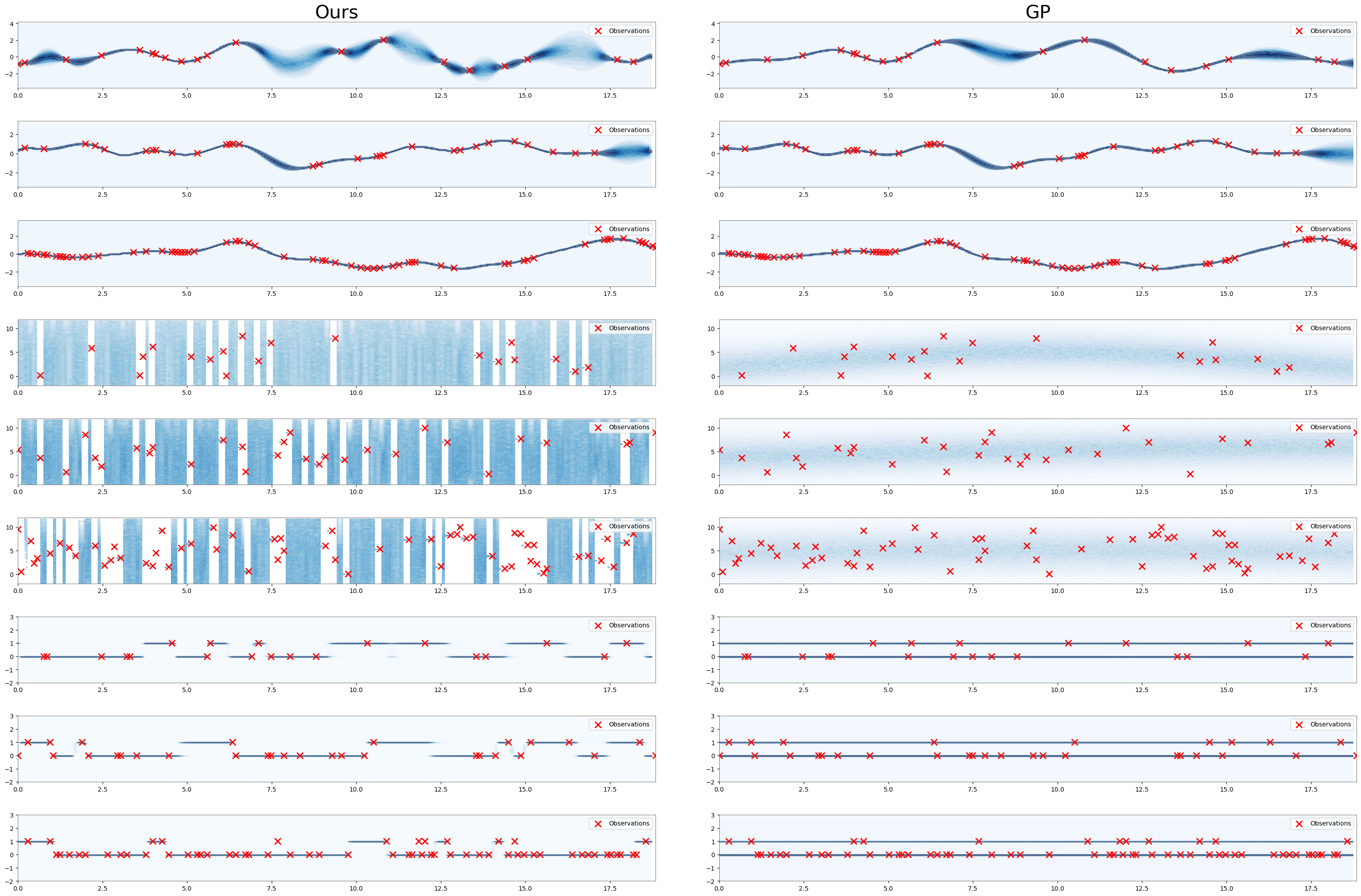}
  \caption{1-D Stochastic Process. We present the modeling results on 1-D data generated from
  the Gaussian Process, Uniform Process, and Markov Process (arranged from top to bottom). For each dataset, observations are set at positions [20, 30, 50]. The red markers indicate the observations,and the blue regions represent the estimating results of the methods. Left: the estimating results of CBC Right: the estimating results of GP.}
  
  \label{fig:1-d results}
\end{figure*}

Generating stochastic process trajectories is a key focus of CBC. We represent the stochastic process as $X=X(\omega,i)_{i\in I} ,\omega \sim P(\omega)$, where $\omega$ is a random variable sampled from a predefined distribution. Using this representation, we can directly construct a simple initial stochastic process, such as a Gaussian white noise process\cite{balakrishnan2011powers}. We define a neural network-based transformation process:$X(i) = Q_\theta(\omega,i)$, where $Q_\theta$ is a neural network parameterized by learnable parameters $\theta$. After sampling entire trajectories from the initial stochastic process, we apply this transformation to obtain trajectories that satisfy observational constraints. We model this transformation process using a Bayesian framework\cite{howson2006scientific}, leveraging Bayes' theorem to infer the conditional distribution.

\begin{equation}
\small P\bigl(X(T)\mid X(S)=O\bigr)= \frac{P\bigl(X(S)=O,\; X(T)\bigr)}{P\bigl(X(S)=O\bigr)}
\end{equation}

The joint probabilities in the numerator can be expressed in an integral form over the initial random variable $\omega$:
\begin{equation}
\small P\bigl(X(S)=O,\,X(T)\bigr)\\
= \int \mathbf{1}\{X(S)=O,\,X(T)\}\,p(w)\,\mathrm{d}w
\end{equation}

The indicator function $\mathbf{1}\{\}$ is:
\[
\mathbf{1}\{X(S)=O,\,X(T)\}
=
\begin{cases}
1, & \text{if X(S)=O}\\[6pt]
0, & \text{otherwise}.
\end{cases}
\]

The marginal probability is likewise obtained via integration, considering only the probability of the observed portion:
\begin{equation}
\small P\bigl(X(S)=O\bigr)\\
= \int \mathbf{1}\{X(S)=O,\,X(T)\}\,p(w)\,\mathrm{d}w
\end{equation}

In practice, directly using an indicator function is not differentiable and thus not well-suited for neural network training. Instead, we adopt a mean squared error (MSE) loss to approximate the constraint 
$X(S)=O$. Specifically, we define
\begin{equation}
\mathcal{L}(\theta)
= \mathbb{E}_{\omega \sim p(\omega)}
\biggl[
  \sum_{s \in S} 
  \bigl(Q_\theta(\omega,s) - O_s\bigr)^2
\biggr].
\end{equation}

By iteratively updating $\theta$ to minimize $\mathcal{L}$ via gradient-based methods, we encourage the network to generate trajectories that closely match the observations at $S$, thereby implicitly satisfying the indicator constraint in a differentiable manner.

During the training phase, we input sampled trajectories from a Gaussian white noise process, and employ a loss function to force the network to output trajectories conditioned on observations. Once trained, we only need to re-input random trajectories, and then the network will generate samples of trajectories that are consistent with observations.

\subsection{FrameWork of Convolution-Based Converter}
The transformation of trajectories between stochastic processes has the potential to estimate $P(X(T)|X(S)=O)$ under a weak prior assumption. However, designing a converter capable of generating stochastic process trajectories with complex dependencies between random variables remains a key challenge.
For instance, a Markov process represents stochastic process structures by enforcing the memoryless property and constructing a state transition matrix, while a Gaussian process employs a covariance function to compute correlation matrices. The effectiveness of these strongly prior-driven representations depends on how well their prior assumptions align with the intrinsic regularities of the problem. However, when the prior is mismatched, it can result in significant performance degradation or even complete failure.
To effectively capture these complex dependencies in stochastic processes, we need an approach based on a weak prior that can adaptively model diverse stochastic processes.
\begin{table*}[t]
\caption{The NLL, for the five methods on the 1-D stochastic process from three different datasets, indicate that CBC, compared to GP, HMM, and WGP, is more adaptive to diverse scenarios and exhibits superior adaptability to limited data compared to DDN. }
\label{1-d-Table}
\begin{center}
\begin{small}
\begin{sc}
\begin{tabular}{c|ccc|ccc|ccc}
\toprule
& \multicolumn{3}{c|}{\textbf{GP$\downarrow$}} & \multicolumn{3}{c|}{\textbf{UP$\downarrow$}} & \multicolumn{3}{c}{\textbf{MP$\downarrow$}}\\
\small \# & 50 & 100 & 150 & 50 & 100 & 150 & 50 & 100 & 150 \\
\midrule

\small Ours &-0.88 &\textbf{-0.95} &-0.9 &\textbf{-0.26} &\textbf{-0.39} &\textbf{-0.6} &-0.65 &-0.69 &\textbf{-0.72}\\
\small GP &\textbf{-0.89}  &-0.92 &\textbf{-0.92} &-0.02 &-0.02 &-0.02 &-0.62 &-0.61 &-0.63\\
\small WGP &-0.88 &-0.92 &-0.92 &-0.02 &-0.02 &-0.02 &-0.05 &-0.05 &-0.05\\
\small HMM  &-0.05 &-0.05 &-0.05 &-0.01 &-0.03 &-0.04 &\textbf{-0.71} &\textbf{-0.72} &-0.71\\
\small DDN   &-0.38 &-0.56 &-0.70 &-0.17 &-0.26 &-0.41 &-0.63 &-0.58 &-0.63\\

\bottomrule
\end{tabular}

\end{sc}
\end{small}
\end{center}

\end{table*}

We interpret the network's goal as mapping an initial random process into one with complex dependencies via trajectory transformations. These complex dependencies manifest in the trajectory generation process and may involve long-range correlations, multi-level structures, non-stationarity, or nonlinear relationships. Convolution stands out as a powerful candidate for building such a converter network, due to its more flexible modeling capacity. In traditional frameworks like Gaussian processes or Markov processes, the prior structure is explicit and closed-form; by contrast, CBC places the prior primarily in the network’s architecture design and parameter initialization, yielding greater adaptability.

Furthermore, the learned kernel (i.e., the convolutional kernel) can be viewed as a “learnable covariance structure” that is no longer bound by strict positive-definiteness or 'memoryless' assumptions, thus accommodating more intricate scenarios. Convolution naturally supports hierarchical feature extraction, enabling the model not only to capture local dependencies but also to progressively incorporate contextual information and ultimately learn longer-range or even global structures.

The network framework is introduced as below, as shown in Figure\ref{fig:model}.

\noindent \textbf{Initial Stochastic Process} \quad  We start with a white noise stochastic process that is independent and has no prior dependency structure, ensuring flexibility in subsequent modeling of dependencies.
\begin{equation}
X(i)=X(w,i),w\sim p(w)
\end{equation}
\noindent \textbf{Preliminary dependency constructor} \quad By employing several layers of MLP mapping\cite{rumelhart1986learning}, we create a preliminary model of relationships between stochastic process trajectories.
\begin{equation}
X^{(0)}(i)=\sigma(b^{(0)}+\sum_{k=0}^{K-1}w_{j,i}X_i+b_j,j=1,...,m)
\end{equation}

\noindent \textbf{Convolution-Based multi-layer dependency constructor} \quad A local-to-global dependency structure is constructed by exploiting multi-layer convolutional \cite{lecun1998gradient} designs with sliding kernels to iteratively combine local dependencies and extend the scope outward.
\begin{equation}
X^{(1)}(i)=\sigma(b^{(1)}+\sum_{k=0}^{K-1}X(i-k)w^{(1)}(k)
\end{equation}
\begin{equation*}
...
\end{equation*}
\begin{equation}
X^{(n)}(i)=\sigma(b^{(n)}+\sum_{k=0}^{K-1}X(i-k)w^{(n)}(k)
\end{equation}

where $K$ denotes the size of the convolutional kernel, $b$ denotes the bias, $X^{(j)}$ is the layer index of the convolution, and $i$ is the index of the random variable.

We have also specifically designed a Smooth Convolution Converter shown as Figture \ref{fig:Deconv} for smooth stochastic processes. It employs Deconvolution consisting of multi-layer upsample-convolution-layers to effectively construct smooth trajectories, thereby modeling smooth stochastic processes in images. This design enhances the model's ability to generate continuous and coherent image completions, as the Smooth Convolution Converter adapts flexibly to the inherent data patterns without being constrained by specific prior structural assumptions.

\noindent \textbf{Output Layer} \quad The final layer maps above output to the trajectory values of the expected stochastic process. If needed, a specific activation function can be applied to map the trajectory values to a designated output space.
\begin{equation}
X(i)=\sigma(X^{(n)}(i))
\end{equation}
\section{Experiment}


\subsection{1-D Stochastic Processes}
In this subsection, we test CBC for modeling 1-D stochastic processes.  VVFWe generate three different toy datasets, each representing a distinct modeling task for stochastic processes in different scenarios: a Gaussian process with an inherently smooth kernel, a Markov process with intrinsic temporal dependencies, and a uniform process with independent samples. The datasets are designed as follows:
\begin{itemize}[leftmargin=1.5em,itemsep=0pt,topsep=0pt]
\item \textbf{Gaussian Process Dateset:}
$GP\sim ( m(i), K(i,i')), \quad$
$m(i) = 0, k(\mathbf{i}, \mathbf{i}') = \sigma^2 \exp \left( -\frac{\|\mathbf{i} - \mathbf{i}'\|^2}{2\ell^2} \right),\sigma = 1.0,\ell=1.0,i\in [0,6\pi]$

\item \textbf{Uniform Process Dataset:} 
$X(i) \overset{\text{i.i.d.}}{\sim} \mathcal{U}(-4, 4),\\ 
 i \in [0,6\pi]$
\item \textbf{Markov Process Dateset:}
$
\mathrm{MP} \sim P(X)
$

\textbf{
$
P(X) = \prod_{i=1}^{N} P(X_i \mid X_{i-1})  \\
= \frac{1}{Z} \exp (- \sum_{i} \theta_i X_i - \sum_{i,j} \theta_{ij} X_i X_j ),X(i) \in \{0,1\}
$
}

\end{itemize}
\begin{figure*}[htbp]
  \centering
  \includegraphics[width=0.8\textwidth]{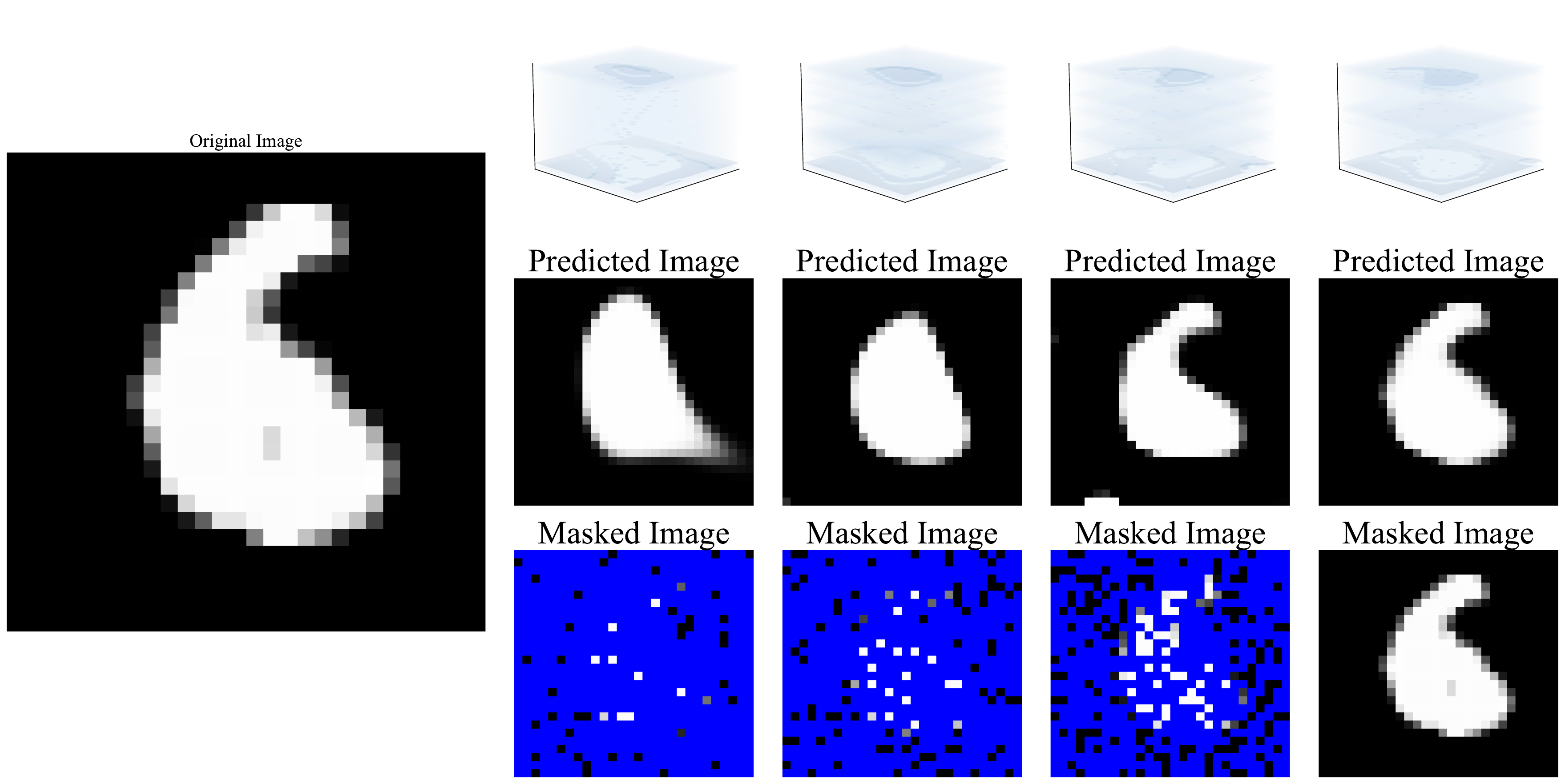}
  \caption{2-D stochastic process on the MNIST dataset. Left: Original Image. Right: From bottom to top, there are images with [100, 200, 500, and 784] observations(blue pixels denote the pixels of targets that have been masked ), respectively, the middle row displays sampled images estimated by CBC, and the top row shows the conditional probability distribution estimated by CBC. As the number of observations increases, the conditional probability estimation becomes more precise.}
  \label{fig:MNIST}
\end{figure*}

We sampled three sequences from each specific random process (where three denotes the number of test sequences), with each sequence comprising 200 random variables. Subsequently, we applied masking at arbitrary positions within these sequences. To evaluate the model's uncertainty inference capabilities under both low-sample and ample-sample conditions, we incrementally increased the number of masked points, specifically setting the number of masked points to [50, 100, 150].

Using the CBC model described in Figure\ref{fig:model}, we sample from a Gaussian white noise process with a sequence length of 5, which is then fed into a three-layer MLPs. The output of the linear layer is reshaped into a multi-channel representation. We then apply 5 layers of transposed convolution to generate the final conditional estimation results. Finally, we impose constraints at observations X(S)=O and optimize the model via gradient descent until the loss converges.

Our experiment compares the performance of the CBC with the Gaussian Process (GP) 
\cite{wilson2011gaussian}, the Hidden Markov Model (HMM) \cite{eddy1996hidden} , Wrapped Gaussian Process (WGP) \cite{lazaro2012bayesian}, and the neural network–based CDE method Deconvolutional Density Network (DDN) \cite{chen2022deconvolutional} across different types of datasets, using negative log-likelihood (NLL) as the evaluation metric in Table \ref{1-d-Table}.

On datasets generated by Gaussian processes and Markov processes, GP and HMM demonstrate superior performance due to their priors being highly aligned with the respective data generation processes. However, when the priors are not suitable, such as applying GP to the Markov process datasets and HMM to the Gaussian process datasets, both methods fail. In contrast, the CBC method, although it does not achieve outstanding results on either type of dataset, maintains performance comparable to scenarios where the priors are well-suited for GP and HMM, and does not completely fail even when the priors are unsuitable.

On the dataset generated by uniform processes, methods such as GP, WGP, and HMM exhibit significant limitations due to their fixed prior assumptions, making it difficult to effectively model the uniform process with independent variables. Although CBC also faces the challenge of accurately capturing data patterns, its strong generalization ability allows the probability distribution of the targets to cover a broader range, encompassing more possibilities. As a result, CBC performs better than the methods relying on strong prior assumptions.

We compared the CBC method with DDN, the results indicate that DDN's performance is highly dependent on the number of training data. In scenarios with limited data, DDN struggles to model effectively, and HMM faces similar challenges. However, as the amount of data increases,the performance of DDN gradually improves, though even with a sample size of 150, its performance ceiling remains comparable to other methods. CBC exhibits a similar trend, but it is noteworthy that in limited data scenarios, CBC clearly outperforms DDN, highlighting the advantage of the CBC method in such scenarios.

The experimental results demonstrate that CBC offers superior flexibility and adaptability, enabling the modeling of diverse stochastic processes. As the number of observations increases, the accuracy of the model estimates continues to improve. Additionally, in scenarios with limited data, our method exhibits strong generalization capabilities compared to other approaches.
\subsection{2-D Stochastic Processes}
We consider image completion as a problem of modeling a 2-D stochastic 
process\cite{garnelo2018conditional}. Specifically, we consider each pixel in an image as a random variable within this process, where its value is inherently dependent on its two-dimensional spatial indices. The spatial dependencies and the underlying image structure dictate that this stochastic process evolves across both dimensions. Consequently, a model designed for image completion must be adapted to capture these two-dimensional dependencies. To this end, we replace the 1-D Convolution-Converter with 2-D Convolution-Converter, allowing the model to operate directly within the two-dimensional image space. Additionally, we employ a 2-D initial Gaussian white noise process to provide a suitable initial state for the image generation process. A sigmoid activation layer was used as the final layer to constrain the output pixel values within the range of [0, 1]. 

When training, we first randomly select an image from the dataset. Subsequently, we select a subset of pixels from the selected image to serve as observations. We input an initial two-dimensional Gaussian white noise vector, and the model then outputs a 2-D stochastic process trajectory, representing the pixel values across the entire image space predicted based on the given observations.

\begin{table}[]
\caption{Pixel-wise NLL for all of the pixels in the image completion task on the MNIST dataset of selected the top 10 images with increasing number of observations [100, 300, 500, 784] and the observations are chosen randomly. With any number of observations CBC outperforms GP and WGP.}
\label{MNIST-table}
\begin{center}
\begin{small}
\begin{sc}
\begin{tabular}{lcccr}
\toprule
\small  & 100$\downarrow$ & 300$\downarrow$ & 500$\downarrow$ & 784$\downarrow$\\
\midrule
Ours & \textbf{-1.9388} & \textbf{-2.9691} & \textbf{-3.3306} & \textbf{-3.6191} \\ 
GP & 0.8303 & 0.8250 & 0.8210 & 0.8211 \\ 
WGP & 0.8291 & 0.8267 & 0.8265 & 0.8263 \\  
\bottomrule
\end{tabular}
\end{sc}
\end{small}
\end{center}
\end{table}
We evaluate CBC on the MNIST dataset and CIFAR dataset. We input a 2-D initial Gaussian white noise sampled vector, which pass through 3 1-D Convolution layers with kernel size = 1, and then reshape to appropriate sizes of 7x7 (for MNIST) and 4x4 (for CIFAR), through 3 Deconvolution layers, and finally output the trajectories(pixels values of images).

We assess the model's performance using the NLL of the ground truth pixels of the original image under the modeled stochastic process's conditional probability distribution. As shown in the Figure \ref{fig:MNIST} and Table \ref{MNIST-table}, CBC can reasonably infer the conditional density distribution of trajectories based on observations. When the number of observations is small, the model's inference uncertainty is high, but as the number of observation points increases, the model's prediction for targets becomes progressively refined, and the completed image increasingly approximates the original image.

We compared our method with existing baseline approaches in Table \ref{MNIST-table}. The results demonstrate that GP and WGP methods are fundamentally limited by the Gaussian distribution assumption, leading to blurred images. In contrast, CBC demonstrates superior adaptability to data patterns, yielding more plausible prediction outcomes.

\begin{table}[]
\caption{Pixel-wise NLL was evaluated for all pixels in the image completion task on the CIFAR-10 dataset. We selected the top 10 images, each with an increasing number of randomly chosen observations [100, 300, 500, 1024]. As the number of observed pixels increases, the performance of the CBC method improves.}
\label{CIFAR-table}
\begin{center}
\begin{small}
\begin{sc}
\begin{tabular}{lcccr}
\toprule
\small  & 100$\downarrow$ & 300$\downarrow$ & 500$\downarrow$ & 1024$\downarrow$\\
\midrule
Ours & 1.0265 & 0.8751 & \textbf{0.7361} & \textbf{0.2216} \\ 
GP & 0.8040 & \textbf{0.7956} & 0.7906 & 0.7847 \\ 
WGP & \textbf{0.8054} & 0.8003 & 0.8026 & 0.8044 \\ 
\bottomrule
\end{tabular}
\end{sc}
\end{small}
\end{center}
\end{table}

\section{Conclusion}
This paper introduces Convolution-Based Converter (CBC), a novel method for modeling stochastic processes that addresses the limitations of traditional SDEs, Markov models, and Gaussian Processes, as well as limited-data challenges for neural network-based conditional density estimation methods. By leveraging a weak-prior assumption and employing a Convolution-Based converter to transform initial stochastic process trajectories into observation-constrained expected trajectories, CBC adaptively learns dependencies without relying on pre-defined stochastic process structures, thus enhancing flexibility and adaptability. 
\section{Acknowledgements}
This work was supported by National Natural Science Foundation of China under Grant No. 61872419, No. 62072213. Shandong Provincial Natural Science Foundation No. ZR2022JQ30, No. ZR2022ZD01, No. ZR2023LZH015. Taishan Scholars Program of Shandong Province, China, under Grant No. tsqn201812077. “New 20 Rules for University” Program of Jinan City under Grant No. 2021GXRC077. Key Research Project of Quancheng Laboratory, China under Grant No. QCLZD202303. Research Project of Provincial Laboratory of Shandong, China under Grant No. SYS202201.

\nocite{langley00}

\bibliography{example_paper}
\bibliographystyle{icml2025}




\end{document}